%% file: main.tex
\documentclass{article}
\usepackage{ijcai26}

\usepackage{times}
\usepackage{soul}
\usepackage{url}
\usepackage[hidelinks]{hyperref}
\usepackage[utf8]{inputenc}
\usepackage[small]{caption}
\usepackage{graphicx}
\usepackage{amsfonts}
\usepackage{amsmath}
\usepackage{algorithmic}
\usepackage{amssymb}
\usepackage{amsthm}
\usepackage{booktabs}
\usepackage{algorithm}
\usepackage{algorithmic}
\usepackage[switch]{lineno}

\usepackage{xcolor}
\usepackage{arydshln}
\usepackage{multirow}
\usepackage{booktabs}
\usepackage{makecell}

\title{PACE: A Personalized Adaptive Curriculum Engine for 9-1-1 Call-taker Training}

\author{
Zirong Chen
\and
Hongchao Zhang
\and
Meiyi Ma
\affiliations
College of Connected Computing, Vanderbilt University, Nashville, Tennessee 37235, USA
\emails
\{zirong.chen, hongchao.zhang, meiyi.ma\}@vanderbilt.edu
}

\begin{document}

\maketitle

\begin{abstract}

9-1-1 call-taking training requires mastery of over a thousand interdependent skills, covering diverse incident types and protocol-specific nuances. A nationwide labor shortage is already straining training capacity, but effective instruction still demands that trainers tailor objectives to each trainee's evolving competencies. This personalization burden is one that current practice cannot scale.
Partnering with Metro Nashville Department of Emergency Communications (MNDEC), we propose PACE (Personalized Adaptive Curriculum Engine), a co-pilot system that augments trainer decision-making by (1) maintaining probabilistic beliefs over trainee skill states, (2) modeling individual learning and forgetting dynamics, and (3) recommending training scenarios that balance acquisition of new competencies with retention of existing ones.
PACE propagates evidence over a structured skill graph to accelerate diagnostic coverage and applies contextual bandits to select scenarios that target gaps the trainee is prepared to address. 
Empirical results show that PACE achieves 19.50\% faster time-to-competence and 10.95\% higher terminal mastery compared to state-of-the-art frameworks. Co-pilot studies with practicing training officers further demonstrate a 95.45\% alignment rate between PACE's and experts' pedagogical judgments on real-world cases. Under estimation, PACE cuts turnaround time to merely 34 seconds from 11.58 minutes, up to 95.08\% reduction.

\end{abstract}

\input{1_introduction}
\input{2_motivation}
\input{3_overview}
\input{4_method}
\input{5_eval}
\input{6_related}
\input{7_summary}



\bibliographystyle{named}
\bibliography{ijcai26}

\end{document}

%% file: 1_introduction.tex
\section{Introduction}


9-1-1 call-takers are first points of contact in life-threatening situations, and their guidance makes differences between life and death. Training them for this role is notoriously difficult~\cite{chen2025sim911}. A car crash, for instance, may begin as a routine law enforcement matter requiring only traffic blockage and vehicle description inquiries, but escalates to paramedic involvement if injuries are reported; further if unknown fluid leakage is present, fire-related safety checks become necessary as the fluid might be flammable or even explosive.
Traditional training relies on human trainers who review past performance and select subsequent training objectives accordingly~\cite{logidebrief}. As trainee cohorts grow and experienced trainers remain scarce, this model struggles to scale.
In addition, trainees also differ in how they learn: some acquire skills quickly but forget without reinforcement, while others need repeated exposure but retain knowledge longer~\cite{radvansky2022memory}. Tailoring instruction to these differences improves outcomes substantially, but adds burden for trainers already monitoring progress across many trainees. As a result, most programs default to uniform curricula, overlooking individual learning patterns that could accelerate or hinder competency development.

Despite recent advancements in personalized learning systems, we identify following \textbf{challenges}:
(1) \textit{Engagement-Learning Discrepancy.} Educational recommendation systems often inherit objectives from information retrieval and e-commerce, such as maximizing click-through rates or session duration~\cite{covington2016deep,koren2009matrix}, which do not necessarily align with learning goals. While recent work incorporates learning signals~\cite{verbert2012context}, fundamental limitations remain. Trainees may ``prefer'' scenarios they can complete easily, but skill acquisition requires practice near the edge of competence. In consequences, call-takers may perform well in training simulations yet remain unprepared for high-stress real-world calls involving unfamiliar incident combinations.
(2) \textit{Inductive Knowledge Tracing.} Most knowledge tracing methods~\cite{piech2015deep,ghosh2020akt} estimate mastery from observed performance but typically treat skills as independent. In 9-1-1 call-taker training, over 1,000 distinct skills are involved, making direct assessment of each one infeasible within realistic training horizons. Inductively inferring mastery across structurally related skills is therefore critical for scalable training.
(3) \textit{Fine-Grained Tutoring.} Existing personalized tutoring systems often operate at course, module, or topic granularity~\cite{verbert2012context}, lumping together heterogeneous skills within broad categories. A trainee labeled ``competent in medical calls'' may handle routine chest pain assessments but struggle with overdose intervention or pediatric emergencies. In 9-1-1 operations, such granularity is dangerous. Because dispatching a call-taker based on coarse competence can delay or mis-specify instructions, with potentially fatal consequences.
(4) \textit{Persistent Learner Modeling.} LLM-based tutoring systems show promise for educational dialogue~\cite{wang2025genmentor,gao2025agent4edu} but struggle with longitudinal training~\cite{fatemi2025test}. A typical 9-1-1 program spans 6 to 8 weeks with over 200 simulated calls per trainee; encoding complete interaction histories exceeds practical context limits~\cite{vodrahalli2024michelangelo}. Retrieval quality also degrades when relevant information is buried mid-context~\cite{liu2024lost}. Without persistent temporal modeling, curricula fail to schedule retention practice, and previously mastered skills can erode unnoticed until a failure surfaces weeks after training.

Our \textbf{technical innovations} and \textbf{contributions} are:
(1) We formalize 9-1-1 call-taker training as curriculum optimization over a structured graph.
(2) We propose \textit{PACE}\footnote{Code and details: \url{https://github.com/AICPS-Lab/PACE-911}.}, first-of-its-kind curriculum optimization system designed for 9-1-1 call-taker training that maintains probabilistic beliefs over trainee competence, estimates individual learning dynamics, and selects training scenarios based on defined tutoring context with bandits.
(3) We comprehensively evaluate PACE's performance against state-of-the-art frameworks in terms of time-to-competence and terminal mastery.
(4) We conduct co-pilot studies with experienced training officers at a partnering 9-1-1 center, evaluating PACE's alignment with expert pedagogical judgment under real-world cases. 

Besides technical innovations, PACE delivers following \textbf{social impacts}:
(1) As a co-pilot, PACE's scenario recommendations achieve 95.45\% agreement with experienced training officers' selections, validating that the learned policy captures domain-relevant pedagogical principles rather than exploiting simulation artifacts.
(2) Under estimation, PACE cuts 11.58 minutes turnaround time by 95.08\% to solely 34 seconds, saving up to 11.01 minutes per call and at least 4,756.32 minutes per typical training class.

\noindent \textbf{\textit{Practitioner Collaboration.}} PACE is developed in partnership with local 9-1-1 call center serving a mid-size U.S. city. Domain experts from training, quality assurance, and operations contributed throughout: validating motivating observations against national and local contexts, co-specifying the skill graph structure and annotation schema, reviewing PACE design choices, and participating in the co-pilot evaluation.


%% file: 2_motivation.tex
\section{Motivating Study}
\label{sec:motivation}

To ground our motivation in operational reality, we analyzed the call-taking training manual provided by local 9-1-1 call center, along with 923 training session logs comprising conversational transcripts, evaluation rubrics, and trainer feedback annotations. We report two key observations.

\noindent \textbf{Individualized feedback faces bandwidth constraints.}
Effective training relies on tight feedback loops in which trainees complete a simulated call, receive targeted debriefing, and apply corrections in subsequent attempts. Our analysis shows that the average turnaround time for a single simulation-debriefing cycle is 11.58 minutes, including call review, protocol cross-referencing, error diagnosis, and feedback delivery. With typical cohort structures assigning one trainer to approximately 12 trainees, each completing at least 3 sessions with at least 12 calls per session, full coverage would require over 83 hours of review time per day, which is physically infeasible. This limitation is not specific to any single agency but reflects a structural constraint across the field. The opportunity is clear: assistance that accelerates assessment and surfaces priority issues can help trainers focus their limited time where it is most impactful.

\noindent \textbf{Localized skill gaps cascade into systemic failures.}
We examine how deficiencies in individual skills propagate to overall call-handling performance. Using the procedural manual’s 857 discrete assessment checkpoints organized hierarchically, we generate 500 synthetic call instances with randomized applicable checkpoints and simulate the removal of individual skills. Removing a single skill orphaned, on average, 11.17\% of downstream checkpoints (96 of 857) and caused 48.80\% of complete call evaluations to fail (244 of 500 instances). For example, assessing patient consciousness is a foundational checkpoint shared across cardiac arrest, choking, drowning, overdose, and trauma protocols. This skill gates downstream decisions such as whether to initiate CPR, select between the Heimlich maneuver and recovery position, or assign medical dispatch priority. A trainee who inadequately masters consciousness assessment therefore underperforms across all dependent incident types rather than failing a single protocol. Because training typically progresses through incident types sequentially, such gaps may remain undetected for weeks and compound with forgetting. This interdependent structure motivates curriculum systems that track skills at finer granularity, allowing mastery in one context to inform readiness in others.

%% file: 3_overview.tex
\section{Problem Formulation}
\label{sec:problem-formulation}

We formalize the curriculum design problem as decision-making over a structured skill graph.

\subsection{Knowledge Graph Representation}

Working with on-duty 9-1-1 call-takers and training officers, we annotate and decompose call-taking procedures into fine-grained, assessable competencies and their dependency structure. We encode this structure as a directed knowledge graph, $\mathcal{G} = (\mathcal{V}, \mathcal{E})$. The node set $\mathcal{V}=\mathcal{V}_C \cup \mathcal{V}_Q \cup \mathcal{V}_I$ comprises \textit{condition} nodes $\mathcal{V}_C$ (incident-state premises), \textit{question} nodes $\mathcal{V}_Q$ (information-gathering skills), and \textit{instruction} nodes $\mathcal{V}_I$ (directive-delivery skills). The edge set $\mathcal{E}$ captures: (1) sequential dependencies among questions/instructions (procedural order and prerequisites), (2) implication links from questions/instructions to conditions they establish, and (3) entailment links from conditions to downstream questions/instructions they activate or mandate. The resulting graph contains $|\mathcal{V}| = 1,053$ nodes and $|\mathcal{E}| = 1,283$ edges spanning 63 incident types. 


\subsection{Teaching under Approximation}
\label{sec:pomdp}

Human-centric pedagogical frameworks involve inherently unobservable learner states~\cite{leighton2004attribute,delatorre2009dina}. In high-stakes domains such as emergency call-taking, two trainees may produce identical responses while possessing fundamentally different underlying competencies: one acts from procedural mastery, another guesses correctly under pressure. Trainers cannot directly inspect these latent states and must rely on observable behavior alone. Let $\Theta = \{\theta_v\}_{v \in \mathcal{V}}$ denote the trainee’s true competency state, where each $\theta_v \in [0,1]$ represents mastery of skill $v$. Any curriculum system therefore maintains a belief model $\hat{\Theta} = \{\hat{\theta}_v\}_{v \in \mathcal{V}}$ that approximates $\Theta$ from observations.

\noindent \textbf{Trainer’s Observations.}
After each simulation, structured debriefing produces observations over skills in the activated subgraph. Observations take values $o_v \in \{\top, \bot, \sim, \oslash\}$, where $\top$ and $\bot$ indicate compliance or violation, $\sim$ indicates partial compliance, and $\oslash$ denotes not applicable due to unmet prerequisites. These signals are noisy functions of latent mastery: higher $\theta_v$ increases the likelihood of compliance, but execution variability, stress, and context introduce observation noise. Observations accumulate in a history $\mathcal{H}$, from which beliefs $\hat{\Theta}$ are inferred.

\noindent \textbf{Trainer’s Actions.}
Trainer selects simulations based on the current belief state $\hat{\Theta}$. Each scenario $\mathcal{S} = (\iota, \mathbf{c})$, defined by incident type $\iota \in \mathcal{I}$ (e.g., car crash) and condition configuration $\mathbf{c}$ (i.e., what nodes to activate or deactivate), activates a subgraph $V_{\mathcal{S}} \subseteq \mathcal{V}$. Training proceeds over $N$ sessions. At each session $t$, the trainer selects a batch of $K$ scenarios $\mathbf{S}_t = \{\mathcal{S}_t^{(1)}, \ldots, \mathcal{S}_t^{(K)}\}$ informed by $\hat{\Theta}^{(t)}$. The curriculum design problem is to choose the sequence $\{\mathbf{S}_t\}_{t=1}^{N}$ that maximizes learning outcomes under approximate beliefs.

\paragraph{Objectives.}
Curriculum design balances coupled objectives. The \textit{explicit objective} maximizes terminal competence:
\begin{equation} \max_{\{\mathbf{S}_t\}} \; \mathbb{E}\left[\sum_{v \in \mathcal{V}} \theta_v^{(N)}\right] \label{eq:explicit-objective} \end{equation}
The \textit{implicit objective} minimizes the approximation gap $\delta = \frac{1}{N}\sum_i \|\mathbb{E}[\hat{\theta}_i] - \theta_i\|$, since scenario selection quality degrades as $\hat{\Theta}$ diverges from $\Theta$~\cite{tabibian2019enhancing,su2023optimizing}.

%% file: 4_method.tex
\section{PACE for 9-1-1 Call-taking Training}
\label{sec:methods}

In this section, we present \textit{PACE} (\textbf{P}ersonalized \textbf{A}daptive \textbf{C}urriculum \textbf{E}ngine), firs-of-its-kind system for adaptive curriculum optimization for 9-1-1 call-taking training.

\subsection{System Overview}
\label{sec:system-overview}

\begin{figure}[h]
    \centering
    \includegraphics[width=0.85\linewidth]{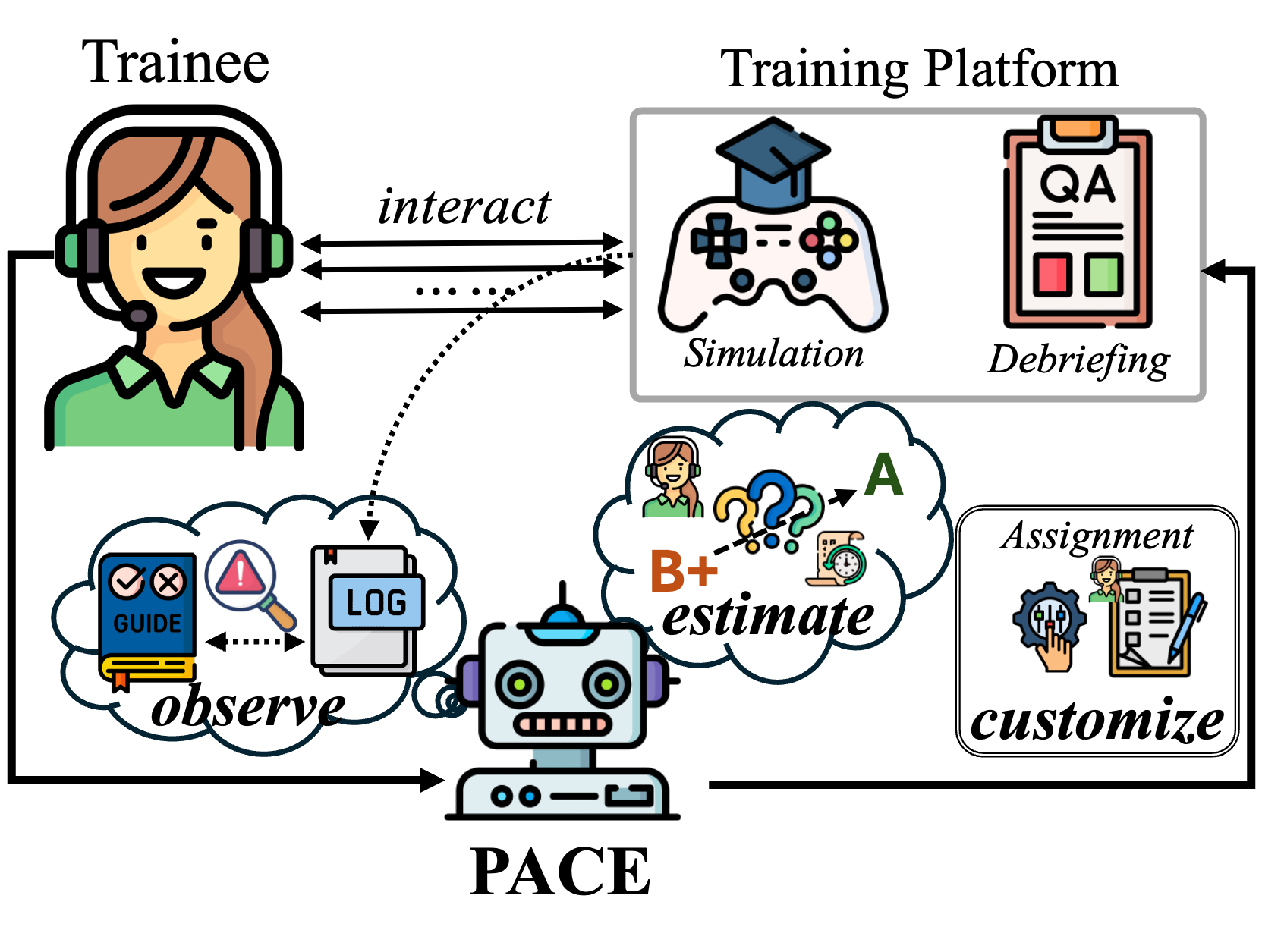}
    \caption{Overview of PACE's involvement in 9-1-1 training.}
    \label{fig:overview}
\end{figure}

In operational settings, PACE operates as a curriculum layer within existing 9-1-1 training infrastructure that role-plays callers and conducts performance debriefing. PACE observes trainee interaction logs, including simulation instantiations, conversation transcripts, and debriefing results, and compares observed learning outcomes against target call-taking knowledge to optimize and personalize each trainee’s learning, in Figure~\ref{fig:overview}.
PACE consists of:
(1) \textit{Belief Tracker}, Section \ref{sec:belief-tracker}, that updates per-node Beta posteriors from transcripts and propagates evidence via similarity;
(2) \textit{Dynamics Estimator}, Section \ref{sec:dynamics-estimator}, that infers trainee-specific learning pace $\lambda$ and forgetting rate $\psi$ using a surrogate model; and
(3) \textit{Contextual Bandit}, Section \ref{sec:contextual-bandit}, that selects scenario batches by balancing exploitation and exploration under the tutoring context.

\subsection{Pre-runtime Preparation: Similarity Index}
\label{sec:similarity}

Skill transfer enables inference over unobserved nodes based on performance on related skills. We precompute a similarity index $\Phi$ offline. For each node $v$, we obtain a semantic embedding $\mathbf{s}_v \in \mathbb{R}^{384}$ by encoding its textual description using a pretrained sentence transformer~\cite{reimers2019sentence}. Pairwise similarity combines semantic relatedness with positional compatibility~\cite{vaswani2017attention}: 
\begin{equation}
\phi(v_i, v_j) = \cos(\mathbf{s}_i, \mathbf{s}_j) \cdot \exp\bigl(-\epsilon |\mathbf{d}_i - \mathbf{d}_j|\bigr)
\label{eq:similarity}
\end{equation}
where $\mathbf{d}_v$ is normalized depth and $\epsilon$ controls position sensitivity. The exponential term reflects that skills at similar protocol stages (e.g., both early information-gathering) transfer more readily than skills at different stages (e.g., initial assessment vs.\ final instructions). Following standard practice in embedding-based retrieval~\cite{li2025matching}, we cache only pairs exceeding a preset threshold to maintain sparsity. 




\begin{figure*}[t]
    \centering
    \includegraphics[width=\textwidth]{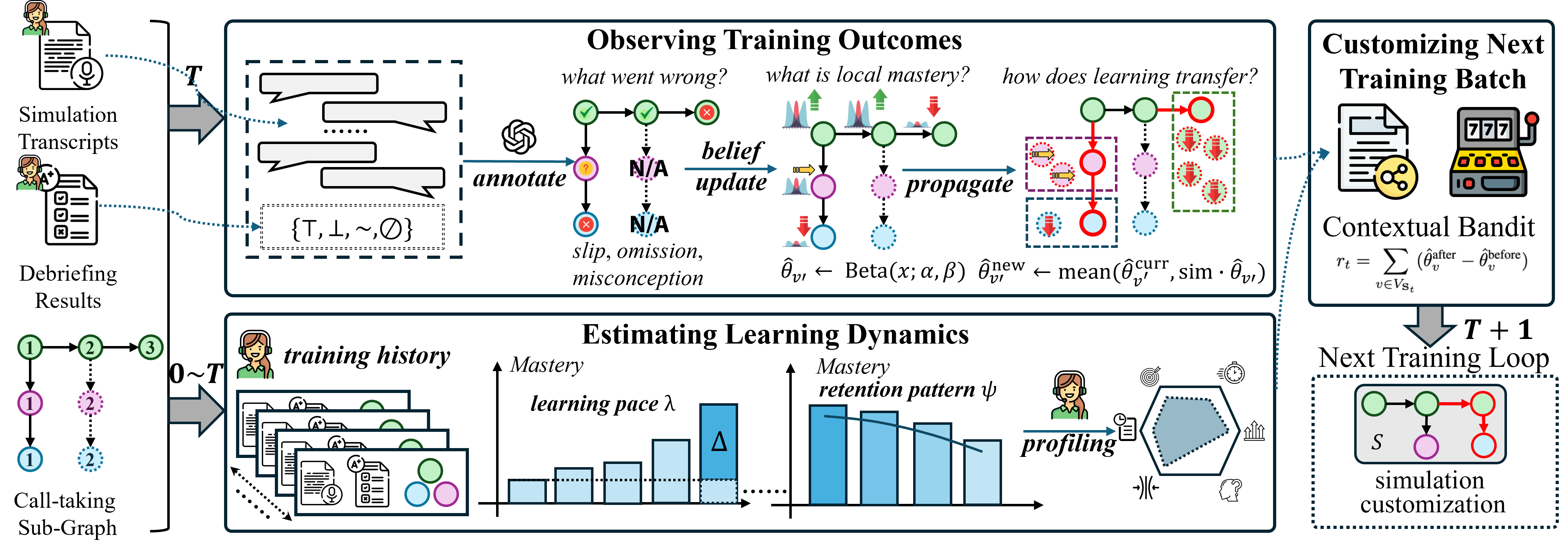}
    \caption{PACE runtime under adaptive phase after cold start.}
    \label{fig:pace}
\end{figure*}

\subsection{PACE Runtime}

The complete PACE runtime proceeds in two phases: 
(1) \textit{Cold Start Phase}: During $t = 1, \ldots, N_0$, PACE silently observes trainee performance on externally-assigned scenarios without making recommendations. This phase accumulates observations to initialize beliefs and estimate learning dynamics. We set $N_0 = 15$ based on local training logistics; and (2) \textit{Adaptive Phase}, see Figure \ref{fig:pace}: From session $N_0 + 1$ onward, PACE actively selects batches with contextual bandit after observing learning outcomes and estimating learning dynamics. 

\subsection{Observing the Training Outcomes}
\label{sec:belief-tracker}

The Belief Tracker maintains $\hat{\Theta}$ through observation extraction and belief updates.

\noindent \textbf{Observation Extraction.}
An LLM extracts structured observations from call simulation transcripts~\cite{he2024harnessing}. For each node $v \in V_{\mathcal{S}}$, the system outputs: outcome $o_v \in \{\top, \bot, \sim, \oslash\}$; error type $e_v \in \{\textsc{slip}, \textsc{misconception}, \textsc{omission}\}$~\cite{liu2023advances}; and a prompted flag indicating whether the caller prompted the action versus independent performance.

\noindent \textbf{Belief Updates.}
For each node $v$, we maintain a Beta posterior $\mathrm{Beta}(\alpha_v, \beta_v)$ representing uncertainty over mastery $\theta_v$~\cite{shen2024survey}. The posterior mean $\mu_v = \alpha_v / (\alpha_v + \beta_v)$ estimates mastery, while the variance $\sigma_v^2$ captures epistemic uncertainty.
Given an observation $o_v \in \{\top, \bot, \sim\}$, the posterior is updated using weighted pseudo-counts:
$\alpha_v \leftarrow \alpha_v + w^{+} \mathbf{1}\{o_v \in \{\top, \sim\}\}$
and
$\beta_v \leftarrow \beta_v + w^{-} \mathbf{1}\{o_v \in \{\bot, \sim\}\}$,
where $\mathbf{1}\{\cdot\}$ denotes the indicator function and $w^{+}, w^{-} > 0$ control the strength of positive and negative evidence.
Weights are modulated by error type~\cite{pelanek2017bayesian}: prompted successes receive $w^+ = 0.5$; misconceptions receive $w^- = 1.5$; slips receive $w^- = 0.5$. 

\noindent \textbf{Inter-Node Mastery Propagation.} When node $v$ is directly observed with given mastery, we inductively update evidence to similar nodes: $\hat{\theta}_{v'}^{\text{new}} \leftarrow \text{mean}\bigl(\hat{\theta}_{v'}^{\text{curr}}, \; \phi(v, v') \cdot \hat{\theta}_v\bigr)$. 

\subsection{Estimating the Learning Dynamics}
\label{sec:dynamics-estimator}

Besides widely used practice and cognitive factors~\cite{chen2023disentangling,cheng2024dygkt}, the Dynamics Estimator captures individual differences with a surrogation~\cite{tabibian2019enhancing,ma2023each} focusing on learning (denoted with $\lambda$) and forgetting dynamics (denoted with $\psi$). 

\noindent \textbf{Learning Pace $\lambda$.}
We operationalize $\lambda$ as the average marginal mastery gain per practice opportunity over the trainee's interaction history $\mathcal{H}$:
\begin{equation}
    \lambda = \frac{1}{|\mathcal{H}|} \sum_{(v,t) \in \mathcal{H}} \Delta\theta_v^{(t)}
\end{equation}
where $\Delta\theta_v^{(t)}:=\theta_v^{(t)} -\theta_v^{(t-1)}$ denotes mastery updates for skill $v$ at practice session $t$. Trainees with higher $\lambda$ reliably consolidate new skills with fewer repetitions, enabling faster progression to complex or compositional scenarios.

\noindent \textbf{Forgetting Model $\psi$.}
Skills decay over real time $\tau$ following a power law, one dominant model in cognitive science~\cite{su2023optimizing}.
Let $\Delta \tau_v$ denote the elapsed clocked time since skill $v$ was last practiced.
We model mastery decay as:
\begin{equation}
    \theta_v^{(\tau + \Delta \tau_v)} = \theta_v^{(\tau)}\cdot (1 + \kappa \cdot \Delta \tau_v)^{-\psi}
    \label{eq:forgetting}
\end{equation}
where $\psi>0$ is a trainee-specific forgetting rate and $\kappa>0$ is a constant that normalizes elapsed time into the effective scale at which forgetting accumulates.
The power-law structure captures the empirically observed phenomenon that forgetting slows over time: recently acquired skills decay more rapidly than skills that have undergone consolidation.
Modeling decay over real time rather than session count is operationally important in training settings.
We estimate $\psi$ by fitting Eq.~\eqref{eq:forgetting} to observed retention outcomes when previously mastered skills are reassessed after variable temporal gaps.

\subsection{Customizing the Next Training Batch}
\label{sec:contextual-bandit}

PACE formulates scenario selection as a contextual bandit problem~\cite{belfer2022raising,nguyen2024generating}, where a policy maps trainee-specific contexts to scenario choices that maximize expected learning gain under uncertainty. 

\noindent \textbf{Context Vector.}
At session $t$, we construct a compact context vector that summarizes the current training state:
\begin{equation}
    \mathbf{x}_t = [\underbrace{\bar{\sigma}^2, c}_{\text{model quality}}, \underbrace{\hat{\lambda}, \hat{\psi}, \bar{w}, d}_{\text{trainee state}}, \underbrace{t/N}_{\text{progress}}]
    \label{eq:bandit}
\end{equation}
where $\bar{\sigma}^2$ is the mean belief variance over skill mastery (direct observation of $\Theta$ is not available under deployment, we use $\bar{\sigma}^2$ as one quality indicator of the surrogate model from trainer's side), $c$ is current coverage, $\hat{\lambda}$ is estimated learning pace, $\hat{\psi}$ is estimated forgetting rate, $\bar{w} \in \mathbb{R}$ is the average mastery of weak skills (under preset threshold), $d \in \mathbb{R}$ is the number of skills near the forgetting threshold, and $t / N$ denotes normalized training progress over $N$ total sessions. 

\noindent \textbf{Action Space.}
Each action corresponds to a scenario $\mathcal{S} = (\iota, \mathbf{c})$, defined by an incident type $\iota$ and contextual configuration $\mathbf{c}$. 297 candidate scenarios are filtered to satisfy prerequisite, operational validity, and complexity constraints.
PACE selects a batch of $K=5$ scenarios sequentially, updating the context after each selection to reflect newly covered skills. 

\noindent \textbf{Reward.}
The reward reflects observed learning gain induced by the selected scenario after applying real time decay:
\begin{equation}
    r_t = \sum_{v \in V_{\mathcal{S}_t}} \left( \hat{\theta}_v^{(t)} - \hat{\theta}_v^{(t-1)} \right)
\end{equation}
where $V_{\mathcal{S}_t}$ is the activated subgraph by $\mathcal{S}_t$. Measures of model uncertainty appear in the context rather than in the reward, allowing the bandit to learn when exploratory actions that reduce uncertainty lead to higher downstream gains.

\noindent \textbf{Algorithm.}
We adopt Thompson Sampling~\cite{agrawal2013thompson}, maintaining posterior distributions over expected rewards and sampling from these posteriors to select actions. This approach naturally balances exploration and exploitation under uncertainty. We model the expected reward as $\mathbb{E}[r \mid \mathbf{x}, \mathcal{S}] = \boldsymbol{B}_{\mathcal{S}}^\top \mathbf{x}$, where $\boldsymbol{B}_{\mathcal{S}}$ denotes scenario-specific coefficients with a Gaussian prior. Parameters are updated via Bayesian linear regression after observing rewards.


%% file: 5_eval.tex
\section{Evaluation}

We evaluate PACE via 
(1) \textit{system-level comparisons} with ablations, 
(2) \textit{within-PACE analysis} across tutoring granularities and behavioral patterns, 
and (3) a \textit{co-pilot study} with domain experts under operational conditions. 
All experiments run on a machine with 128,GB RAM, an AMD Ryzen Threadripper Pro 7975WX CPU, and an NVIDIA RTX 6000 Ada GPU. 

\subsection{Role-played Trainee Agents}
Although this study has received IRB approval, \textit{direct evaluation with human trainees is not ideal for controlled curriculum analysis} due to:
(1) training programs span multiple weeks, limiting rapid iteration;
(2) trainee competency states are latent and unobservable, precluding reliable ground-truth labels; and
(3) curriculum refinement requires counterfactual comparisons that are infeasible because instructional exposure irreversibly alters knowledge states.
Following recent work on LLM-based educational simulation~\cite{zhang-etal-2025-simulating,liu2024personality}, we construct role-played trainee agents with controllable ground-truth parameters. Each agent
$\mathcal{A} = (\Theta, \lambda, \psi, \mathcal{P}, \mathcal{M})$
consists of a ground-truth competency state
$\Theta = \{\theta_v\}_{v \in \mathcal{V}}$, where $\theta_v \in [0,1]$ denotes mastery of skill $v$,
learning and forgetting rates $(\lambda, \psi) \in \mathbb{R}^2$,
a behavioral persona $\mathcal{P}$ encoding response patterns,
and a scratchpad $\mathcal{M}$ that maintains learning history across sessions~\cite{park2023generative}.
Given a training scenario $\mathcal{S}$ that activates a skill subset $V_{\mathcal{S}}$, the agent generates utterances by sampling: $u \sim \text{LLM}(\mathcal{P}, \mathcal{S}, \mathcal{M}, \{(\theta_v, e_v) : v \in V_{\mathcal{S}}\})$,
where $e_v$ denotes an error type sampled with probability $1 - \theta_v$.
To prevent knowledge leakage, prompts explicitly constrain the agent to act as a trainee with specified skill levels~\cite{zhao2025entityflow_parametric_contextual,wang2024rolellm}.
Ecological validity is further improved by incorporating error templates extracted from anonymized training logs~\cite{dong2024survey}.
We define four trainee archetypes parameterized by $(\lambda, \psi)$:
\textit{Fast Learner} $(0.12, 0.15)$ with rapid acquisition and slow decay;
\textit{Moderate Learner} $(0.07, 0.25)$ reflecting typical progression;
\textit{Struggling Learner} $(0.03, 0.35)$ with slow acquisition and rapid decay; and
\textit{Quick Forgetter} $(0.10, 0.45)$ with strong initial acquisition but rapid decay.
The parameters $(\lambda, \psi)$ are hidden from PACE during evaluation, and only conversational observations are available.
Each archetype includes 10 instantiations with $\pm 15\%$ parameter noise.
We use OpenAI GPT-5 for PACE and all re-implemented agentic baselines.
To avoid in-family comparisons, trainee agents are instantiated using Claude Sonnet 4.5.

\input{_main_table}

\subsection{System-level Comparison with Ablation}
\label{sec:system-level}

\noindent \textbf{Goal.}
This experiment configuration studies \textbf{\textit{does PACE help trainee learn effectively and efficiently?}}

\noindent \textbf{Baselines.}
We compare against:
(1) \textit{Round-Robin}, sequentially cycling through incident types;
(2) \textit{Deficit-Driven}, deterministically selecting scenarios targeting lowest-mastery skills based on debriefing;
(3) \textit{GraphRAG+LLM}~\cite{edge2024graphrag}, which we selected as the strongest vanilla LLM integration after pilot trials, using graph-guided retrieval and traversal to contextualize deficits prior to selection; and
(4) agentic pipelines\footnote{All agentic baselines are re-implemented from published descriptions to operate on the 9-1-1 call-taking data used in this study.} including GenMentor~\cite{wang2025genmentor} and Agent4Edu~\cite{gao2025agent4edu}.

\noindent \textbf{Metrics.}
(1) \textit{Coverage@$t$} measures the mastery cardinality fraction by session: 
$\text{Coverage}@t = |\{v : \theta_v^{(t)} \ge 0.85\}| / |\mathcal{V}|$;
(2) \textit{Zero-to-Hero} (Z2H) is the number of sessions required to first reach episode-level competence (scenario score $\ge 0.85$), capturing learning efficiency;
(3) \textit{Random Exam} (RE) reports performance (0--100) on a held-out exam uniformly sampled across incident types, simulating certification-style slot-filling assessment used in local training practice.
For each method, we train 40 simulated trainees for 50 sessions with batch size $K=5$. Every 5 sessions, we insert a 24-hour gap to reflect realistic training schedules.

\noindent \textbf{Ablation variants.}
(1) PACE \textit{w/o propagation} removes similarity-based belief transfer over the skill graph;
(2) PACE \textit{w/o dynamics} replaces trainee-specific parameters $(\lambda_i,\psi_i)$ with fixed population averages $(\bar{\lambda}=0.08,\bar{\psi}=0.30)$.

\noindent \textbf{Analysis.} From Table~\ref{tab:main-results}, on Fast Learners, PACE reaches Z2H of 22.19 compared to Agent4Edu at 27.58, corresponding to 19.50\% reduction, while achieving higher coverage (C@50: 95.27\% vs.\ 88.18\%). The performance gap widens for more challenging archetypes. On Quick Forgetter ($\psi=0.45$), PACE attains C@50 of 91.22\%, whereas GenMentor degrades from 71.66\% at session 10 to 67.51\% at session 50, indicating negative learning progress. This regression reflects GenMentor struggles to handle high forgetting rates without explicit temporal modeling. Agent4Edu maintains positive progress but plateaus below PACE (C@50: 86.74\% vs.\ 91.22\%), as its static learner memory does not adapt to inter-session decay.
Ablation results confirm complementary contributions of the two components. Removing propagation substantially increases Z2H across archetypes (e.g., Fast Learner: 46.13 vs.\ 22.19), since the system must directly observe each skill rather than inferring mastery via graph-based transfer. Removing dynamics estimation disproportionately harms archetypes that deviate from population norms. For Quick Forgetter, RE drops from 91.04 to 80.16 ($\Delta$10.88) because the population-average $\bar{\psi}=0.30$ underestimates the true $\psi=0.45$, leading to insufficient retention scheduling. With both components enabled, propagation accelerates belief coverage while dynamics estimation calibrates beliefs to individual learning trajectories. Modeling learning as a temporal process rather than a static profile is therefore essential under realistic session gaps. In summary, \textit{with all design choices empirically validated, PACE outperforms baseline training approaches in both effectiveness and efficiency.}

\subsection{Within-PACE Analysis}

We observe and report PACE's performance on different tutoring \textit{granularity levels} and its overall \textit{behavioral patterns}.

\subsubsection{Granularity Analysis of PACE}

\noindent \textbf{Goal.}
This section studies \textbf{\textit{whether fine-grained belief tracking in PACE improves learning outcomes}}. While prior work and motivating studies suggest benefits of finer-granularity tutoring, we seek empirical evidence under 9-1-1 training.

\noindent \textbf{Granularity Levels.}
We compare three variants:
(1) \textit{PACE-coarse}, which tracks beliefs at the departmental level (police, fire, medical; 3 categories);
(2) \textit{PACE-medium}, which tracks beliefs at the incident-type level (e.g., car crash, structure fire; 63 categories); and
(3) \textit{PACE-fine}, which tracks beliefs at the skill level (1,053 nodes) and corresponds to the full PACE system used in prior experiments.
At coarser granularities, beliefs are aggregated and scenarios are selected based on aggregate deficits.
All variants are evaluated on the same four trainee archetypes over 50 sessions.

\begin{figure}[h]
    \centering
    \includegraphics[width=0.95\linewidth]{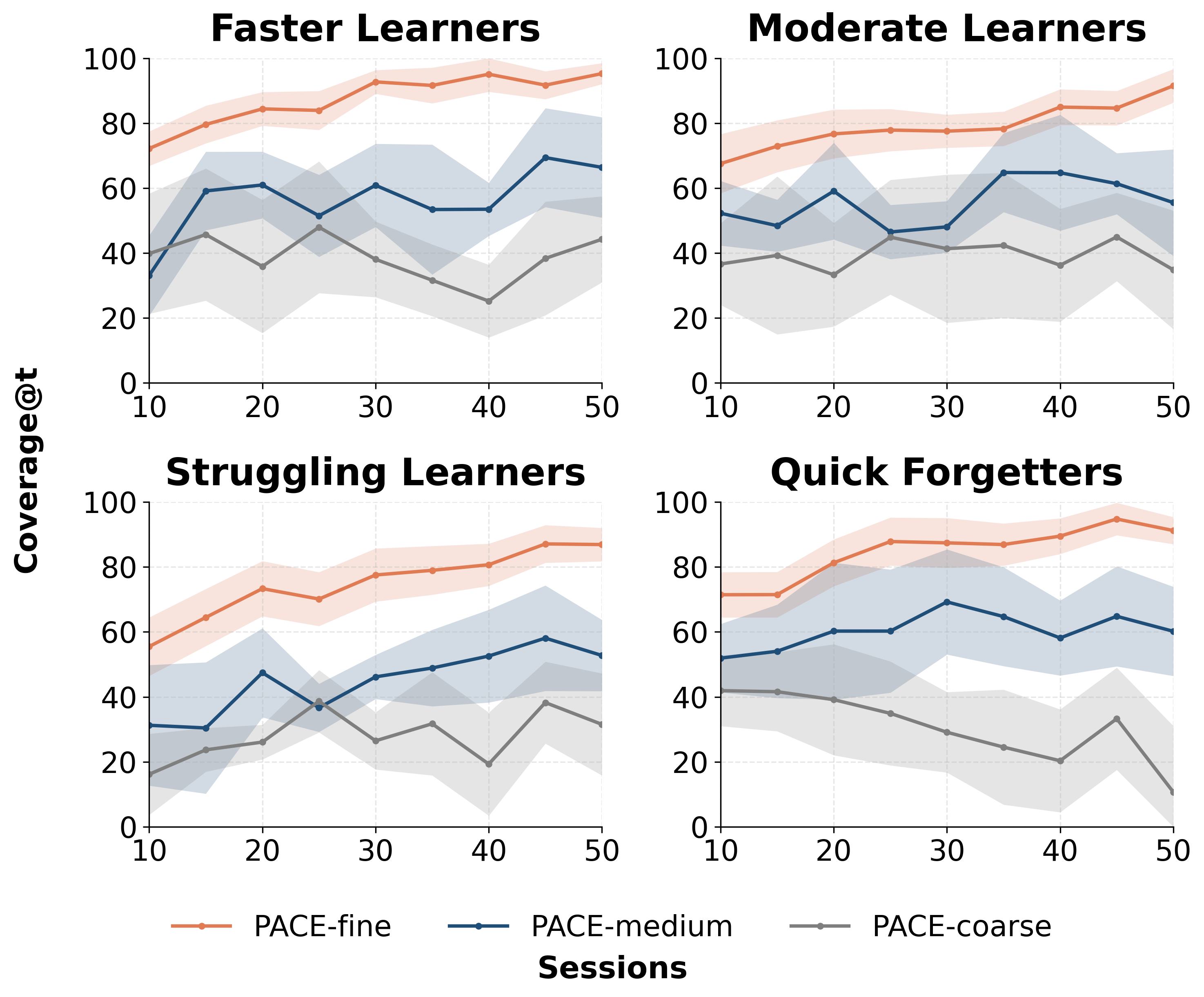}
    \caption{Coverage progression under different belief granularities.}
    \label{fig:granularity}
\end{figure}

\noindent \textbf{Analysis.}
Figure~\ref{fig:granularity} shows consistent advantages for fine-grained belief tracking across all trainee archetypes. PACE-fine achieves terminal coverage (C@50) of 95.27\%, 91.51\%, 86.91\%, and 91.22\% for the four archetypes, respectively, compared to 68.49\%, 56.11\%, 50.93\%, and 59.67\% for PACE-medium, with PACE-coarse performing worst overall. PACE-medium exhibits intermediate performance but with higher variance, particularly for the Struggling and Quick Forgetter archetypes. This instability arises because incident-type granularity conflates heterogeneous skills within each category. For example, a trainee may perform well on basic choking response yet struggle with choking complications, which type-level tracking cannot distinguish. Overall, these results show that \textit{fine-grained belief tracking in PACE is not merely a design choice but a requirement for reliable curriculum optimization in high-stakes 9-1-1 training.}


\subsubsection{Behavioral Analysis of PACE}
\noindent \textbf{Goal.} This section studies \textbf{\textit{does PACE correctly approximate trainee learning states and take expected actions on the fly?}}

\noindent \textbf{Surfaced PACE Actions.}
We compare PACE’s estimated trainee mastery $\hat{\Theta}$, represented as Beta distributions over 1,053 skill nodes, against the trainee agent’s ground-truth mastery
$\Theta \in [0,1]^{|\mathcal{V}|}$.
We quantify the approximation gap using $\delta$, defined in Section~\ref{sec:pomdp}, which measures the mean absolute deviation between believed and true mastery.
We also report PACE’s internal belief quality indicator
$\bar{\sigma}^2 = \frac{1}{N}\sum_i \mathrm{Var}(\hat{\theta}_i)$,
which directly influences scenario selection.
Each scatter point corresponds to one of 10 independent runs per archetype over 50 sessions.
Point color indicates the explore to exploit ratio within the selected batch for that session.

\begin{figure}
    \centering
    \includegraphics[width=\linewidth]{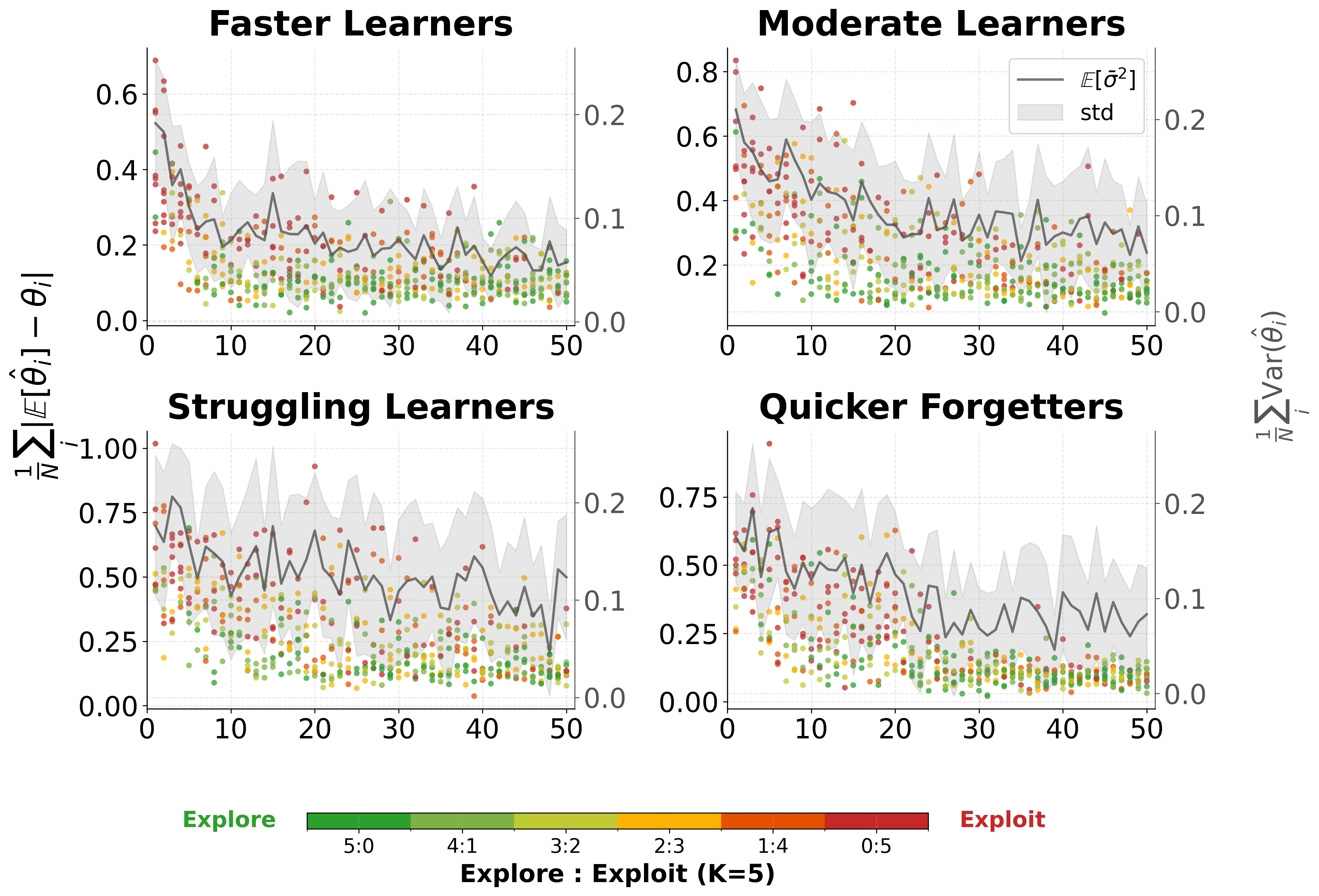}
    \caption{Convergence of approximation gap $\delta$ (scatter points, left axis) and belief variance $\bar{\sigma}^2$ (gray line $\pm$ std, right axis) across sessions. Point color indicates explore:exploit ratio within each batch ($K$=5): green denotes exploration-heavy sessions, red denotes exploitation-heavy sessions.}
    \label{fig:bandit}
\end{figure}

\noindent \textbf{Analysis.}
Figure~\ref{fig:bandit} reveals two key patterns:
(1) the approximation gap $\delta$ decreases across all trainee archetypes, indicating that PACE’s belief estimates converge toward ground truth. Convergence speed varies with trainee complexity: Fast Learners stabilize around session 20, Moderate Learners around session 30, and Struggling Learners around session 40; and
(2) the explore to exploit balance adapts to belief quality. Early sessions exhibit predominantly red and orange points, corresponding to higher $\bar{\sigma}^2$ values. As beliefs consolidate and $\bar{\sigma}^2$ decreases, points shift toward green, indicating increased exploration.
These trends show that PACE translates estimation uncertainty into appropriate action selection. When beliefs are unreliable, PACE prioritizes exploitation to calibrate suspected weaknesses. As belief confidence increases, PACE shifts toward exploration to expand coverage over uncertain regions of the skill graph. Overall, \textit{PACE balances skill acquisition and belief calibration through uncertainty-aware curriculum control.}

\subsection{Case Study: PACE as a Co-Pilot}

\noindent \textbf{Goal.} This section studies \textbf{\textit{how well does PACE align with expert judgment and assist real-world training?}}

\noindent \textbf{Setup.}
We evaluate PACE under operational conditions at our partnering 9-1-1 call center using both quantitative alignment analysis and qualitative expert assessment.
For quantitative evaluation, we collected 923 training catalog entries from recent training classes, each containing simulation logs, debriefing results, and trainer-provided comments with next-session assignments based on trainee performance.
We executed PACE on this dataset and measured alignment between PACE’s curriculum recommendations and expert pedagogical decisions.
For qualitative evaluation, we administered an anonymized survey to 21 domain experts, including 12 training officers and 9 active call-takers with training experience.
Participants rated agreement with PACE’s design choices, such as skill-level diagnostic granularity, and overall helpfulness for call-taker training on a 5-point Likert scale from 1 (strongly disagree) to 5 (strongly agree).

\noindent \textbf{Analysis.}
PACE achieves 95.45\% alignment (881/923) with expert judgment, validating its effectiveness as a training co-pilot under practical usage.
Moreover, PACE reduces average turnaround time from 11.58 minutes to 34 seconds (adaptive phase average), saving 95.08\% of training officer time.
Survey results further support PACE’s design and utility, with participants rating agreement with design choices at 4.62 and overall helpfulness at 4.43.
Open-ended responses highlight two recurring themes.
First, participants emphasized the value of fine-grained diagnosis: ``\textit{Knowing exactly which part of the protocol they missed matters more than just saying they struggled with medical calls}.''
Second, participants noted reduced cognitive burden: ``\textit{It would save me from being overwhelmed by the entire training class}.''
Overall, the co-pilot study confirms that \textit{PACE’s recommendations align with expert pedagogical judgment in real-world settings while reducing training turnaround time by up to 95\%.}

%% file: _main_table.tex
\begin{table*}[t]
\centering
\footnotesize
\resizebox{\textwidth}{!}{
\begin{tabular}{llcccccccccc}
\toprule
& 
& \multicolumn{5}{c}{\textbf{Fast Learner} ($\lambda$=0.12, $\psi$=0.15)}
& \multicolumn{5}{c}{\textbf{Moderate Learner} ($\lambda$=0.07, $\psi$=0.25)} \\
\cmidrule(lr){3-7} \cmidrule(lr){8-12}

\textbf{Group} & \textbf{Method}
& C@10 $\uparrow$ & C@30 $\uparrow$ & C@50 $\uparrow$ & Z2H $\downarrow$ & RE $\uparrow$
& C@10 $\uparrow$ & C@30 $\uparrow$ & C@50 $\uparrow$ & Z2H $\downarrow$ & RE $\uparrow$ \\
\midrule

\multirow{3}{*}{Heuristics}

& Round-Robin    & 42.24$\pm$7.16 & 53.43$\pm$5.57 & 66.11$\pm$7.37 & $\diagup$ & 63.14$\pm$4.11  & 34.19$\pm$8.21 & 47.21$\pm$7.20 & 55.15$\pm$8.16 & $\diagup$ & 51.43$\pm$6.44 \\
& Deficit-Driven & 57.40$\pm$4.14 & 55.22$\pm$6.84 & 56.67$\pm$8.17 & $\diagup$ & 59.46$\pm$6.41  & 49.50$\pm$5.15 & 50.11$\pm$7.13 & 48.00$\pm$8.77 & $\diagup$ & 45.47$\pm$7.17 \\
\midrule

\multirow{3}{*}{LLM-based}
& GraphRAG   & 59.56$\pm$7.72 & 60.50$\pm$8.18 & 61.44$\pm$9.31 & $\diagup$ & 60.22$\pm$9.21  & 51.43$\pm$7.82 & 48.91$\pm$9.22 & 50.46$\pm$8.91 & $\diagup$ & 49.84$\pm$9.64 \\
& GenMentor  & 65.67$\pm$9.55 & 79.41$\pm$8.33 & 84.15$\pm$8.08 & 29.40$\pm$6.60 & 86.87$\pm$6.71  & 66.14$\pm$10.11 & 73.34$\pm$8.78 & 75.44$\pm$6.41 & $\diagup$ & 78.12$\pm$8.83 \\
& Agent4Edu  & 68.14$\pm$8.27 & 82.06$\pm$6.58 & 88.18$\pm$5.57 & 27.58$\pm$5.24 & 90.47$\pm$5.12  & 62.41$\pm$8.10 & 72.59$\pm$9.66 & 84.57$\pm$6.18 & 48.10$\pm$1.90 & 86.17$\pm$5.21 \\
\midrule

\multirow{3}{*}{\textbf{PACE}}

& w/o prop   & 66.74$\pm$7.40 & 81.43$\pm$6.18 & 84.73$\pm$5.19 & 46.13$\pm$3.87 & 86.95$\pm$5.00  & 59.47$\pm$8.20 & 70.25$\pm$6.17 & 83.46$\pm$5.11 & 49.15$\pm$0.85 & 84.71$\pm$6.83 \\
& w/o dyn    & \textbf{72.93$\pm$6.11} & 88.71$\pm$4.23 & 92.00$\pm$4.14 & 34.55$\pm$5.80 & 91.13$\pm$4.06  & 60.15$\pm$7.12 & 70.13$\pm$8.10 & 84.10$\pm$6.17 & 46.10$\pm$3.90 & 85.03$\pm$7.23 \\
& full       & 72.19$\pm$5.31 & \textbf{92.71$\pm$3.65} & \textbf{95.27$\pm$3.23} & \textbf{22.19$\pm$4.11} & \textbf{96.05$\pm$2.54}  & \textbf{67.55$\pm$9.11} & \textbf{77.54$\pm$5.10} & \textbf{91.51$\pm$5.19} & \textbf{34.10$\pm$5.52} & \textbf{90.11$\pm$4.51} \\
\midrule\midrule


&
& \multicolumn{5}{c}{\textbf{Struggling Learner} ($\lambda$=0.03, $\psi$=0.35)}
& \multicolumn{5}{c}{\textbf{Quick Forgetter} ($\lambda$=0.10, $\psi$=0.45)} \\
\cmidrule(lr){3-7} \cmidrule(lr){8-12}

\textbf{Group} & \textbf{Method}
& C@10 $\uparrow$ & C@30 $\uparrow$ & C@50 $\uparrow$ & Z2H $\downarrow$ & RE $\uparrow$
& C@10 $\uparrow$ & C@30 $\uparrow$ & C@50 $\uparrow$ & Z2H $\downarrow$ & RE $\uparrow$ \\
\midrule

\multirow{3}{*}{Heuristics}

& Round-Robin    & 30.41$\pm$5.22 & 44.27$\pm$8.21 & 50.10$\pm$9.81 & $\diagup$ & 53.44$\pm$8.14  & 47.11$\pm$8.93 & 50.12$\pm$7.29 & 51.38$\pm$7.32 & $\diagup$ & 48.19$\pm$7.20 \\
& Deficit-Driven & 37.17$\pm$4.75 & 35.20$\pm$8.91 & 36.44$\pm$7.33 & $\diagup$ & 33.73$\pm$8.13  & 55.17$\pm$5.87 & 50.25$\pm$8.75 & 52.61$\pm$9.51 & $\diagup$ & 51.17$\pm$8.11 \\
\midrule

\multirow{3}{*}{LLM-based}
& GraphRAG   & 35.71$\pm$10.22 & 34.56$\pm$14.80 & 37.07$\pm$9.88 & $\diagup$ & 39.60$\pm$11.36  & 61.08$\pm$9.10 & 57.87$\pm$12.74 & 58.72$\pm$14.15 & $\diagup$ & 59.83$\pm$10.67 \\
& GenMentor  & 44.14$\pm$8.81 & 51.68$\pm$9.37 & 66.17$\pm$10.15 & $\diagup$ & 65.73$\pm$7.45  & \textbf{71.66$\pm$11.37} & 69.43$\pm$9.85 & 67.51$\pm$11.27 & $\diagup$ & 66.37$\pm$9.11\\
& Agent4Edu  & \textbf{57.15$\pm$8.44} & 71.29$\pm$8.73 & 80.48$\pm$8.13 & 47.88$\pm$2.12 & 81.15$\pm$9.19  & 72.54$\pm$8.82 & 77.14$\pm$8.19 & 86.74$\pm$5.51 & 32.55$\pm$6.55 & 85.78$\pm$6.71 \\
\midrule

\multirow{3}{*}{\textbf{PACE}}

& w/o prop   & 47.17$\pm$8.33 & 71.41$\pm$7.12 & 80.47$\pm$5.71 & 49.44$\pm$0.56 & 78.83$\pm$6.22  & 66.17$\pm$8.05 & 76.85$\pm$6.10 & 84.67$\pm$4.11 & 49.50$\pm$0.50 & 86.13$\pm$4.71 \\
& w/o dyn    & 45.15$\pm$10.48 & 68.98$\pm$9.73 & 80.19$\pm$8.65 & 47.13$\pm$2.87 & 82.17$\pm$9.81  & 69.69$\pm$9.14 & 73.43$\pm$8.72 & 82.09$\pm$6.88 & 48.13$\pm$1.87 & 80.16$\pm$8.85 \\
& full       & 55.50$\pm$9.00 & \textbf{77.54$\pm$8.18} & \textbf{86.91$\pm$5.15} & \textbf{44.14$\pm$5.86} & \textbf{88.18$\pm$5.20}  & 71.45$\pm$6.98 & \textbf{87.43$\pm$7.67} & \textbf{91.22$\pm$4.17} & \textbf{25.91$\pm$3.49} & \textbf{91.04$\pm$5.11} \\
\bottomrule
\end{tabular}
}
\caption{System-level comparison with PACE ablations.
C@\textit{t}(\%) denotes coverage at session \textit{t}.
Z2H (zero-to-hero) denotes sessions to first competent episode. Slash means never observed, excluded for average calculation.
RE (random exams) denotes held-out exam score.
}
\label{tab:main-results}
\end{table*}

%% file: 6_related.tex
\section{Related Work}

\textbf{High-Stakes and Emergency Response Training.}
High-stakes training requires deliberate practice near competence boundaries with tight feedback loops. In emergency response, dispatcher performance directly affects survival outcomes~\cite{chen2026real}, motivating simulation-based training. Recent work applies LLM-based simulation to 9-1-1 dispatcher training~\cite{chen2025sim911}, emphasizing realistic caller role-play rather than pedagogical decision-making. 
\textbf{AI for Edu and Educational AI.}
Knowledge tracing models learner states from interaction data~\cite{shen2024survey}, with recent work incorporating graph structure to capture skill dependencies~\cite{cheng2024dygkt}. But these approaches primarily support diagnosis and do not implement integrated diagnose-prescribe tutoring loops. Adaptive learning systems attempt to close this gap using contextual bandits, spaced repetition, and educational recommendation~\cite{belfer2022raising,su2023optimizing}, which correspondingly assume flat skill spaces, over-focus on retaining acquired items, and prioritize engagement over learning. 
Recent LLM-based tutoring agents~\cite{wang2025genmentor,gao2025agent4edu} enable flexible dialogue and learner profiling, but suffers from context rot or overflow as training history enriches. 

%% file: 7_summary.tex
\section{Summary}

This paper introduces PACE, first-of-its-kind co-pilot for adaptive 9-1-1 call-taker training. By integrating inductive belief tracking over structured skill graphs, individual learning dynamics estimation, and contextual bandit-based scenario selection, PACE enables personalized curriculum optimization that accounts for both skill interdependencies and trainee-specific learning trajectories. Experimental results show improvements in time-to-competence and terminal mastery over state-of-the-art baselines. Co-pilot studies with practicing training officers further confirm alignment with expert pedagogical judgment in operational settings.

Locally verified with a 9-1-1 call-taking agency, PACE supports 9-1-1 call centers facing staffing and time constraints by surfacing diagnostic insights and curriculum recommendations that would otherwise require substantial manual effort. With over 6,000 9-1-1 call centers in the US, it potentially scales to broad deployment. As future work, beyond emergency response, PACE framework can extend to other high-stakes procedural training domains, including medical triage, air traffic control, and crisis intervention.

\newpage
\section*{Acknowledgments}

This work was supported in part by the U.S. National Science Foundation under Grants 2427711 and 2443803, the Google Academic Research Award, and the U.S. Department of Education under Grant R305C240010.    
The opinions, findings, conclusions, or recommendations expressed in this material are those of the author(s) and do not necessarily reflect the views of the sponsoring agencies.